# Multi-Stage Training for Abusive Comment Detection in Indic Languages

Pranshu Rastogi
Department of CSE, JIIT Noida
rastogirpranshu29@gmail.com

Madhav Mathur
Department of ICE, NSUT Delhi
madhavmathur2000@gmail.com

Ramaneswaran S
Department of IT, VIT Vellore
s.ramaneswaran2000@gmail.com

Kshitij Mohan
Department of CSE, IIIT Delhi
kshitij19054@iiitd.ac.in

***Abstract*—In recent years social media has become an increasingly popular tool for communication. People use it to share their ideas, exchange information, and discuss thoughts. Given its prevalence and widespread reach, social media must remain a safe space for people. Content generated on social media can be abusive and it has become increasingly important to detect such content. In this paper, we use a language-based preprocessing and an ensemble of several models and analyze their performance of abusive comment detection. Through extensive experimentation, we propose a pipeline that minimizes the false-positive rate (marking non-abusive as abusive) so that these systems can detect abusive comments without undermining the freedom of expression.**

## Introduction

People use social media platforms to express themselves and share information. As social media platforms continue to grow, several problems emerge that need to be solved. One such problem is the identification of abusive content. Abusive content detection is a challenging task. It gets even more challenging in a multilingual setting.

The multi-lingual abusive comment identification challenge attempts to solve this problem. The dataset consists of 700,000 comments taken from the Moj app. The dataset is multi-lingual and includes 15+ Indian languages. The dataset is human-annotated and also provides metadata on explicit feedback such as the number of likes and reports on each comment.

**Challenges**: Detection of abusive comments is an inherently difficult task. Vidgen et al [1] describe in detail the several challenges facing this task that span technical, social, and ethical domains.. The categorization of abusive comments is a social and theoretical task that does not have an objective set of criteria. The annotations to some extent depend on the subjectivity of the annotator. Due to this several abusive content detection datasets have a low inter-annotator agreement resulting in label noise. There are also several linguistic challenges involved in this task, for example, humor or irony, spelling variations, language dependencies, etc make it more difficult to detect abusive content. It's even more challenging to detect abusive comments in Indian languages due to the low resources available for these languages. Indian languages also have regional dialects that are difficult to detect. There is no predefined corpus of abusive words for indic languages, in addition, many comments include a mixture of languages which makes it more difficult to understand the semantics of the comment.

There is a huge gap in the availability of effective pre-trained language models for Indic languages. The available multilingual models for Indic languages are trained on substantially small corpora which makes the models less effective.

To deal with the above scenarios, our approach encompasses five stages of data preprocessing and model training. We first preprocess the given dataset by cleaning and transliterating each comment into its assigned language. The resultant dataset is divided based on the language. In the next stage, We trained several pre-trained Indic language models on the curated dataset. These trained models were used to generate contextual text embeddings. Boosted decision trees were trained using these embeddings. In the next stage, we iteratively use pseudo labeling to allow a model to learn from its predictions. In the final stage, we apply post-processing by assigning thresholds for each individual language to generate our final predictions.

**Contribution :**

(i) Creating an improvised version of the raw original dataset in which the comments are cleaned and transliterated to their assigned language.

(ii) Creating an effective language-wise training pipeline that gives a robust and competitive performance on the given dataset.

## Data & Analysis

About 50% of the comments in the train and test dataset are in Hindi. Hence a major focus was to improve the model's performance for this language. Dravidian languages which include Tamil, Telugu, Malayalam, and Kannada also make up a sizable portion. The length of the comment text ranges from 50 to 300 and about 75% of them are less than 150 in length. So keeping a maximum text length of 150 is a good start to start modeling this data without significant information loss.

The original dataset was in a raw and unclean form. We used some basic preprocessing to remove non-text entities like HTML tags. Another issue that required attention is that comments often had words from other languages or the text was in Latin script. We used the indic-trans library [5] to transliterate the comments to their original language. For example, we transliterated hinglish comments to Hindi. Using this cleaned and transliterated dataset marginally improved our results.

After several experiments, we noticed that our results stagnated. After careful error analysis, we discovered one potential cause of this is label noise in the dataset. In the dataset, there were many instances where an abusive comment would be marked as non-abusive and vice versa.

We manually checked our misclassifications and noticed instances where we feel the annotations are wrong. For example, a sarcastic comment would be marked as abusive and a very toxic comment would be marked as non-abusive. We estimate about 9-10% of the data in the training set have annotation errors.

To test our hypothesis that there is around 9-10% label noise we perform the following experiment. We took all the data samples which were misclassified by one of our models (which had a score of 0.88882), which was about 12% of the training data. In this subset, there were 40000 comments marked as offensive and 24000 marked as non-offensive. We then trained an XLM-R model on this subset and tested it on the dataset and we noticed that it predicted the opposite labels for 98% of the samples. This gave us some crucial evidence for the presence of label noise.

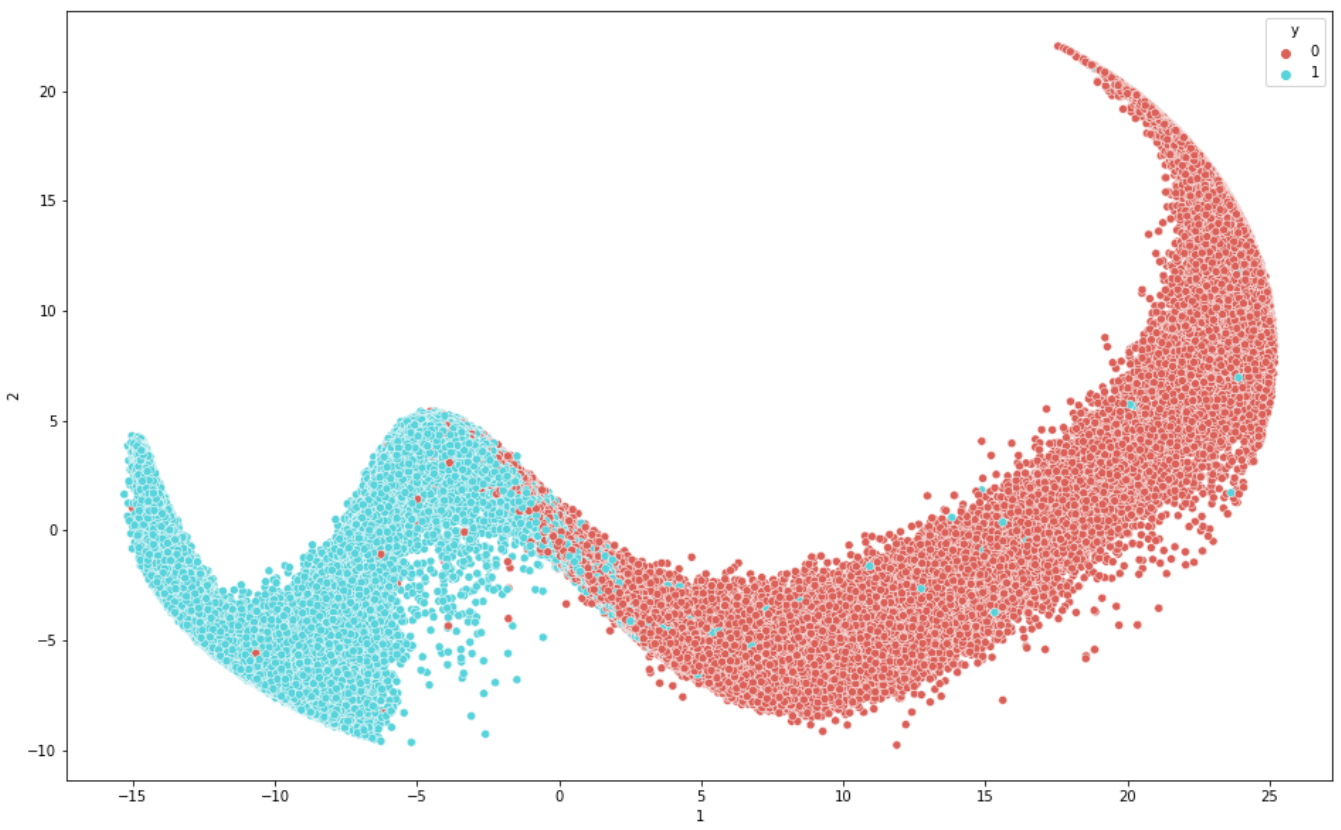

**Fig1.** Dataset Visualization

This notion could be solidified since we know about this noise in the label. We flip all the labels in the training dataset and get a great score of 97% F1 score. but when we tested it on LB we got the same score as we got earlier( 88-89%). This settled our doubt as there is also label noise in the test set.

- In Fig.1 we can see the representation of all training datasets, the dataset is described by taking PCA over the training embeddings and we only converted them from 768 to a 2-dimensional dataset that could be easily plotted with labels provided to us. As we plotted the dataset and noise we came to ensure that that there are 2 clusters one on the left one with Cyan color that cluster is of label offensive and if points get there they are easily defined by our model as non an offensive one, and the one in the red is termed a big cluster of non-abusive dataset samples.

- But when we only plotted the noise in the dataset the distribution of it was very much spread over all of the data. Fig 2. shows us the spread. Where you can see that the labels that were termed as offensive in the actual training dataset are spread over a wide distance and even in very high concentrations of red labels. as now you can infer from both figures. This not only reduces the confidence but also terms as a road blocker for a good F1-Score of 93-95% range.

- Here in Fig 2. we can see the noise which is composed of both the labels. From this figure, we hypothesized that the testing LB will also have an equal proportion of noise with the same distribution over the test samples.

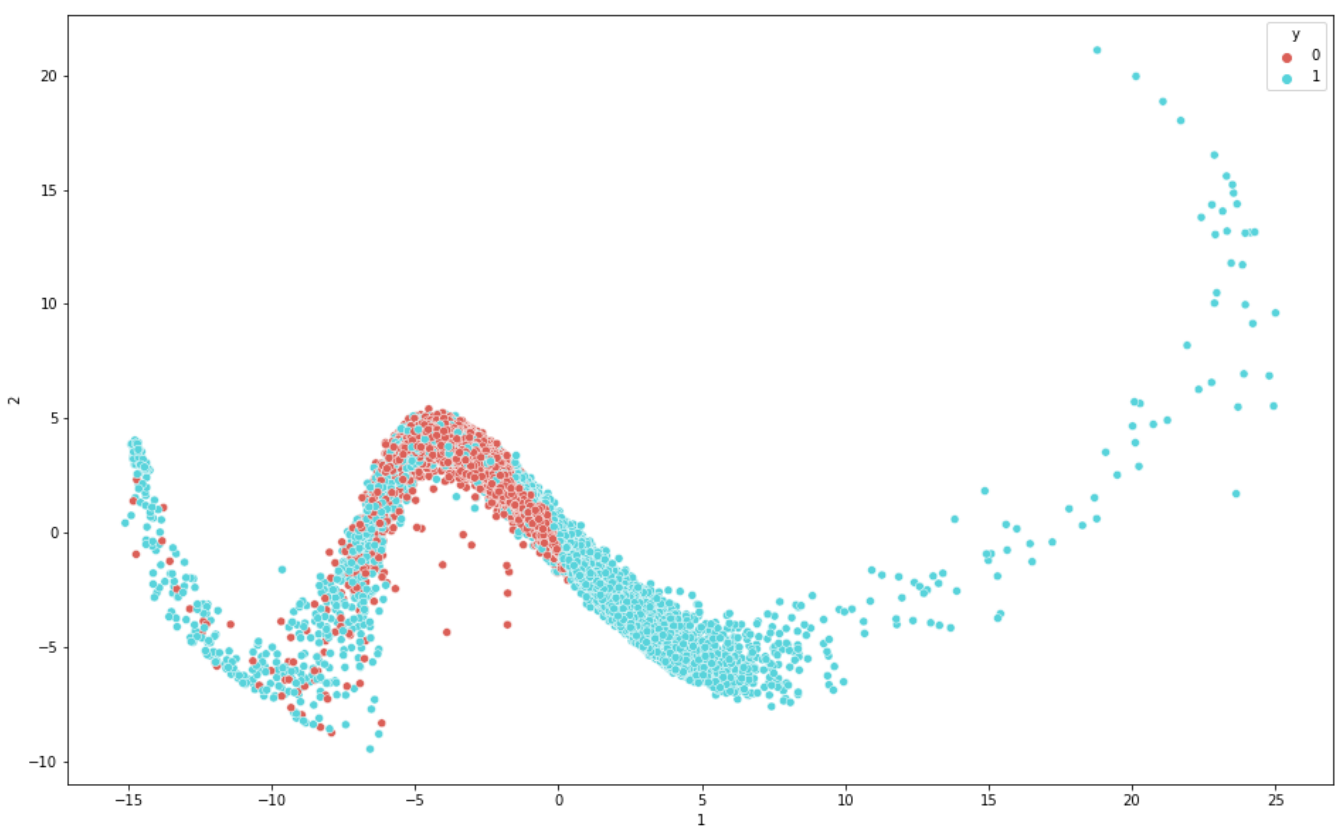

**Fig.2** Noise Distribution

## Proposed Methodology

In this section, we will explain in detail the architecture of our model. We will present our approach in a step-by-step manner, wherein each step we refine our architecture.

### Baseline Approach

For building some baselines we chose Muril [3] and XLM-R [2]. Muril is a BERT model trained specifically for Indic languages. Muril is also trained on transliterated indic data i.e indic data in Latin script since this form of text is prevalent in social media. XLM-R is a Roberta model that is trained on a massive corpus of 100+ languages. We finetune both these models for our task. We notice that XLM-Roberta gave a better CV score of 0.8804 compared to 0.87758 given by Muril. The models mentioned here are the base models

### Explicit Feedback Metadata

The dataset had explicit feedback metadata such as the number of likes and reports. We concatenate these features with the contextual embeddings provided by our base transformer models. We notice a marginal but definite increase in CV score.

### Training on transliterated data

We concatenate the raw comment text and the unclean comment text and pass it as input to our XLM-R model. Using this input, our CV score had a significant increase from 0.8804 to 0.8822. We hypothesize that the combination of raw comment text and the transliterated comment text provides the model with rich syntactic and semantic information to learn the dataset better.

### Further Pretraining On Indic Languages

Since XLM-R is not pre-trained on several of the Indic languages, we attempted to pre-train the model on external datasets for Marathi, Hindi, etc. However, this attempt did not yield any result as we did not notice any improvement in the CV score. This result could be because we were not making any changes in the vocabulary of the model hence newer words were not being picked up leading to no significant learning of the language.

### LGBM Model With XLM-R Embeddings

We trained the XLM-R model mentioned above on 10 folds of the dataset. Then the trained model's pooled output was used as a contextual representation of the text. These contextual vectors concatenated with their corresponding explicit metadata were then used to train an LGBM model. We train the LGBM model on 10 folds and obtained a CV score of 0.89481.

### Oversampling using the clean dataset

We tried an experiment where the original raw dataset and the cleaned dataset were merged to produce a dataset of double the original size. We trained the XLM-R model on this merged dataset with the input to the model being the concatenation of original text and transliterated text, with a maximum text length of 300. This experiment was successful as we were able to improve our CV score from 0.88754 to 0.8882. We then used this fine-tuned model to extract embeddings for LGBM and the LGBM model then gave us a score of 0.89767

### Using TF-IDF Features

We incorporated Tf-Idf embeddings (max length 500) along with the existing XLM-R embeddings and metadata. We then trained an LGBM model on this set of input features on 10 folds, which gave us a score of 0.89772 which is an improvement over the previous best of 0.89767. We then utilized PCA to reduce the dimensions of XLM-R embeddings from 768 to 200. This gave us a minor boost and our score improved to 0.8978

### Language Wise Training Of LGBM

Using the 10 fold embeddings where we scored 0.89767, we train a separate individual LGBM for each language in each fold. This experiment gave us a score of 0.89801 which is a significant boost from our previous score. We also tried language-wise training of XLM-R and Muril which did not yield good results.

### Weighed Ensemble and Post Processing

We performed a weighted ensemble of several model predictions and got a score of 0.8995. We also applied a simple post-processing, in which we selected language-wise thresholds for final predictions. This improved our score to 0.89962 on LB.

**Pseudo Labelling**

We applied the pseudo-labeling technique where we take the predictions of our best-performing model and pass it as input metadata for model training. We then take the predictions of this model and train another model, and this process continues iteratively until the model performance stagnates or worsens. We took predictions of the model which scored 0.89962 for the first iteration of pseudo-labeling and trained the model on 10 folds along with pseudo labels in each fold. We then used these model embeddings along with meta-data to train LGBM on 10 folds as done earlier. This gave us a score of 0.89468 on LB. We also used these embeddings to train language-wise LGBM on 10 folds as done earlier. It performed well and gave a score of 0.89934. This is the boost that we got and led us to get a good model training technique that could produce a score of **0.89934 which is the best score for a single model**. We further incorporated our latest pseudo label wise predictions in our ensemble, which improved our base ensemble score from 0.8995 to 0.89957. After this, post-processing was applied (as done before), which improved our score to **0.90000.**

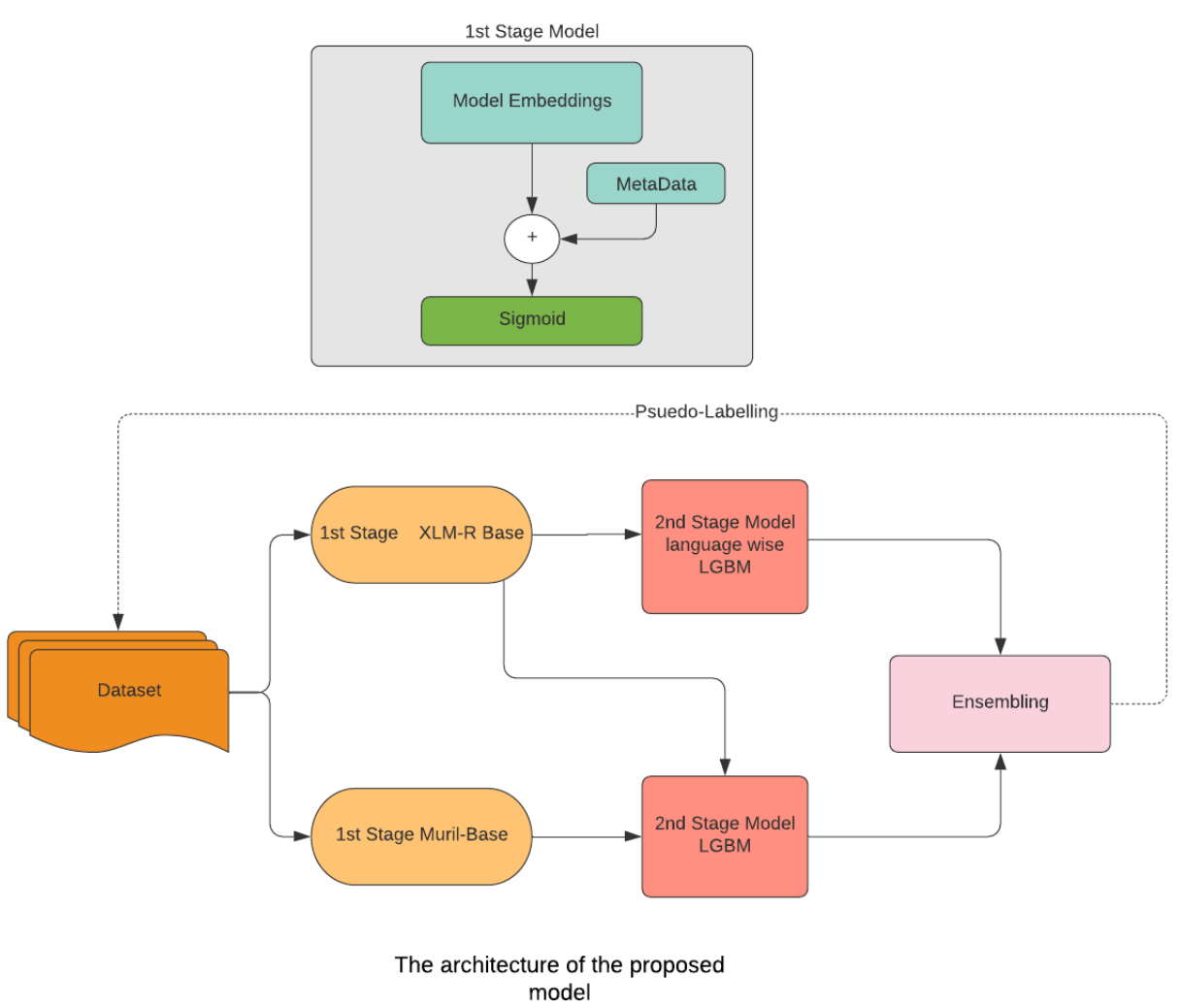

**Figure 3.** Model training pipeline

## Implementation Details

We used an XLM-R base model with a learning rate of 1e-5 and a batch size of 32. Muril had a learning rate of 5e-6 and 32 batch size. We observed that a batch size of 64 leads to less optimal training. We used default LGMB hyperparameters. Strangely, the results with default hyperparameters were better than any hyperparameters search we did, however, a more extensive search could give better parameters. The transformer models were trained using Kaggle TPUs and LGM trained using Kaggle CPUs.

| Model | Test |
|---|---|
| Muril Base (Original Data) | 0.87758 |
| XLM-R Base (Original Data) | 0.88040 |
| Muril Base with Metadata (Original Data) | 0.87820 |
| XLM-R Base 5 Folds (Original Data) | 0.88202 |
| XLM-R Base with Metadata (Cleaned Data) | 0.88754 |
| XLM-R Base with Metadata (Cleaned Data) with LGBM | 0.89481 |
| XLM-R Base with Metadata (Cleaned Data + Original Data) with LGBM | 0.89767 |
| XLM-R Base with Metadata (Cleaned Data + Original Data) with LGBM (Language Wise) | 0.89801 |
| XLM-R Base with Metadata (Cleaned Data + Original Data) Pseudo with submission(89.960) | 0.89468 |
| XLM-R Base with Metadata (Cleaned Data + Original Data) Pseudo with LGBM (Language Wise) | 0.89934 |

**Table 1.** Results obtained from the different architectures

## Results

Our final solution is a weighted ensemble of several high-scoring models which scored 0.89957 on the test set. The models that significantly contributed to the ensemble were XLM-R Base Pseudo with LGBM (0.89468) on 10 Folds and LGBM language-wise with XLM-R Base Pseudo Embeddings on 10 Folds (0.89934). After that, post-processing was applied in which language-wise thresholds were set for predictions. This simple trick improved our score to **0.90000**.

Though we used some innovative modeling procedures there is still a lot that could be done to improve performance. We can optimize the models better by using schedulers. Significant work can be done towards handling the noise in labels.